\documentclass[review]{elsarticle}

\usepackage{amsmath, mathtools, amssymb}
\usepackage{float}
\usepackage{bm}
\usepackage[ruled,vlined]{algorithm2e}
\usepackage{multirow}
\usepackage{soul,color}
\soulregister\cite7
\soulregister\ref7
\soulregister\pageref7
\usepackage{adjustbox}
\usepackage{tikz}
\usetikzlibrary{shapes.geometric, arrows, fit}
\usepackage{lineno,hyperref}
\modulolinenumbers[5]
\usepackage{makecell}

\def\Proof{\noindent{\sl Proof.}\ }
\def\qed{{\hfill $\Box$ \medbreak}}

\tikzstyle{startstop} = [rectangle, rounded corners, minimum width=2cm, minimum height=1cm,text centered, draw=black]
\tikzstyle{io} = [trapezium, trapezium left angle=70, trapezium right angle=110, minimum width=2cm, minimum height=1cm, text centered, draw=black]
\tikzstyle{process} = [rectangle, minimum width=3cm, minimum height=1cm, text centered, text width=3.8cm, draw=black]
\tikzstyle{decision} = [diamond, minimum width=3cm, minimum height=1cm, text centered, draw=black]
\tikzstyle{arrow} = [thick,->,>=stealth]
\tikzset{container/.style={draw, rectangle, dashed, inner sep=.3em}}
\tikzset{container1/.style={draw, rectangle, dashed, inner sep=1em}}

\newtheorem{defi}{Definition}[section]
\newtheorem{thm}[defi]{Theorem}

\journal{Expert Systems With Applications}

\biboptions{authoryear}

\begin{document}

\begin{frontmatter}

\title{A Feature Selection Based on Perturbation Theory}

\author[mymainaddress]{Javad Rahimipour Anaraki}
\ead{jra066@mun.ca}

\author[mysecondaryaddress]{Hamid Usefi\corref{mycorrespondingauthor}}
\cortext[mycorrespondingauthor]{Corresponding author}
\ead{usefi@mun.ca}

\address[mymainaddress]{Department of Computer Science, Memorial University of Newfoundland, \\St. John’s, NL, A1B 3X5 Canada}
\address[mysecondaryaddress]{Department of Mathematics and Statistics, Memorial University of Newfoundland, \\St. John’s, NL, A1C 5S7 Canada}

\begin{abstract}
Consider a supervised dataset $D=[A\mid \textbf{b}]$, where $\textbf{b}$ is the outcome column, rows of $D$ correspond to observations, and columns of $A$ are the features of the dataset. A central problem in machine learning and pattern recognition is to select the most important features from $D$ to be able to predict the outcome. In this paper, we provide a new feature selection method where we use perturbation theory to detect correlations between features. We solve $AX=\textbf{b}$ using the method of least squares and singular value decomposition of $A$. In practical applications, such as in bioinformatics, the number of rows of $A$ (observations) are much less than the number of columns of $A$ (features). So we are dealing with singular matrices with big condition numbers. Although it is known that the solutions of least square problems in singular case are very sensitive to perturbations in $A$, our novel approach in this paper is to prove that the correlations between features can be detected by applying perturbations to $A$. The effectiveness of our method is verified by performing a series of comparisons with conventional and novel feature selection methods in the literature. It is demonstrated that in most situations, our method chooses considerably less number of features while attaining or exceeding the accuracy of the other methods.
\end{abstract}


\end{frontmatter}


\section{Introduction}
In machine learning and pattern recognition, feature selection is the process of selecting the most important features of a problem while removing unnecessary ones. This process plays an important role in reducing the dimension of datasets. Feature selection methods are categorized into two main groups of feature ranking and feature subset selection  \citep{Hall}. The former is a set of methods that ranks the features based on some measured values, and selects the top features, accordingly. The latter screens the critical features using fitness value. Both groups can be implemented using filter-based or wrapper-based approaches  \citep{Kohavi}. In the filter-based approach, a merit evaluates the quality of every feature regardless of its impact on the outcome, while the wrapper-based approaches measure the effectiveness of the features based on the results of a (a set of) classifier(s). The wrapper-based methods are highly computationally-intensive and powerful in predicting the outcome compared to the filter-based methods which are faster but less accurate.

With the emergence of high dimensional data, for example in Genomics, sophisticated feature selection methods are required to remove noisy features and  detect correlation between features. It is desired that a small subset of features are selected to predict the outcome with high accuracy. The traditional feature selection methods such as principal component analysis \citep{PCA} or Relief \citep{Kira} have shortcomings in terms of dimensionality reduction, accuracy, as well as running time. We shall review some of the breakthrough  methods that are effective in these respects.

There have been  numerous methods based on the information theory, see for example \cite{Zhao,SunXin,Bennasar}. These methods aim to minimize the feature redundancy while maximizing the features' relevancy. Most notable and widely used information theory based method is  minimal-redundancy-maximal-relevance criterion (mRMR) \cite{Peng}. It is shown in various studies that  mRMR effectively chooses a small subset of features to predict the outcome with high accuracy. However, as it is pointed out in \citep{Yamada2018}, the computational cost of mRMR on large dataset is high. In  other words, it is not feasible to scale up mRMR for big datasets.

Feature selection is also referred to as variable selection in Statistics. Fundamental variable selection methods include least absolute shrinkage and selection operator (LASSO) and least angle regression (LARS). LASSO, introduced by Tibshirani \citep{Tibshirani}, is a subset selection based on least squares regression. It minimizes the size of a regression model by removing those predictor variables with zero-valued coefficients by calculating Equation \ref{LASSO}, the LASSO estimate, subject to $\sum_{j} |\beta_j| \le t$, where $\beta$ is a vector of coefficients and  $t \ge 0$ is tuning parameter 

\begin{equation}\label{LASSO}
	(\hat{\alpha}, \hat{\beta}) = \arg \min \left\{ \sum_{i=1}^{N} \left\{b_i - \alpha - \sum_{j} \beta_jx_{ij} \right\}^2  \right\},
\end{equation}
and the solution for $\alpha$ is $\hat{\alpha} = \bar{b}$, $\hat{\beta} = (\hat{\beta}_1, \dots, \hat{\beta}_n)^T$ are LASSO estimates where $n$ is the total number of features, $\textbf{b}$ represents responses, $x$ contains predictor variables and $N$ is the number of samples.

LARS, introduced by Efron et al.  \citep{Efron}, is a linear regression model fitting based on the LASSO algorithm which calculates all the LASSO estimates efficiently, in combination with a forward stage-wise linear regression method within $n$ steps, where $n$ is number of covariates and $m$ is number of samples. LARS starts with selecting the most relevant feature and continues by adding the next feature with the highest correlation with the current residual. Then, it continues in a direction which has equal angle from the two already selected features until the next feature is met. The complexity of LARS algorithm is $O(n^3 + mn^2).$

In a novel work, Yamada et al. \citep{yamada2014high} proposed a non-linear feature selection method for high-dimensional datasets called Hilbert-Schmidt independence criterion least absolute shrinkage and selection operator (HSIC-Lasso), in which the most informative non-redundant features are selected using a set of kernel functions, where the solutions are found by solving a LASSO problem. The complexity of the original Hilbert-Schmidt feature selection (HSFS) is $O(n^4)$. In a recent work  \citep{Yamada2018} called Least Angle Nonlinear Distributed (LAND), the authors have improved the computational power of the HSIC-Lasso. They have demonstrated via some experiments that   LAND and HSIC-Lasso have  attain similar classification accuracies and  dimension reduction. However, LAND has the advantage that it can   be deployed on parallel distributed computing.

A method proposed by Chen et al. \citep{chen2017semi} is a feature selection called rescaled linear square regression (RLSR), where a set of coefficients for least square regression is employed to scale and rank features. The advantage of their method is that it can be applied to both supervised and semi-supervised classification problems.

In this paper, we introduce a new linear feature selection method. Linear models usually outperform nonlinear models over high-dimensional datasets. Consider a dataset $D$, consisting of $m$ samples where each sample contains $n+1$ features. Let us denote by $A$ the first $n$ columns of $D$ and by $\textbf{b}$ the last column. Our objective is to remove those columns of $A$ that do not have a significant impact on $\textbf{b}$. So, we want to choose a subset of columns of $A$ to express (up to an error) $\textbf{b}$ as a linear combination of this subset. We consider the linear system $AX=\textbf{b}$, where $X=[x_1, \ldots, x_n]^T$ is the vector of unknowns. In practical applications, the system $AX=\textbf{b}$ may not have exact solutions. However, we want to find an $X$ so that the distance between $AX$ and $\textbf{b}$ is as small as possible. That is, we want to minimize the distance $||AX - \textbf{b}||_2$ over all $X$. To do so, we shall use the method of least squares and singular value decomposition (SVD) of $A$. The Moore-Penrose inverse $A^{+}$ of $A$ is defined in terms of SVD of $A$ and it is known that $X=A^{+}\textbf{b}$ is the unique solution with the smallest 2-norm that satisfy the least square problem $\text{min}_{X} ||AX - \textbf{b}||_2$, see Theorem \ref{ls-solutions}.

There has been extensive literature, see \citep{Golub}, regarding the sensitivity of solutions of least square problems when $A$ is full-rank. It is also known and rightfully cautioned that solutions of singular systems where condition number of $A$ is bigger than one are sensitive to perturbations in $A$. However, we prove in Theorem \ref{main-thm}, that one can use perturbations to reveal correlations between columns of $A$. To do so, we solve both $AX=\textbf{b}$ and $(A+E)\tilde X=\textbf{b}$ using SVD, where $E$ is a small perturbation of $A$. It turns out that features $\textbf{f}_i$ and $\textbf{f}_j$ correlate if and only if  $\mid x_i-\tilde x_i\mid$ and $\mid x_j-\tilde x_j\mid$ are close (in the magnitude of $||E||_2$.). This allows to cluster features based on the differences $\mid x_i-\tilde x_i\mid$. 

Next, we consider the column vector $|X-\tilde X|$ whose values are $\mid x_i-\tilde{x_i}\mid$ and consider  clustering features based on this single column. As we mentioned, features that correlate with each other fall into the same cluster. However, within a cluster there might be features that do not correlate (but have the same value for $\mid x_i-\tilde{x_i}\mid$). To break down some big clusters that  contain independent features,  we use a simple but efficient method based on the angle between features. 
In Section \ref{refining clustering}, we consider the projection of $\textbf{b}$ into each of the hyperplanes obtained by removing one feature at a time. We construct a column that consists of the angles between each feature and the corresponding hyperplane. The third column in our clustering process consists of angles between each feature and $\textbf{b}$. 

We note that often in classification problems and real-world datasets, for example Cancer datasets, the column $\mathbf{b}$ contains  nominal values (classes). One can then assign numerical values for each class. Although, this assignment is not unique our method is insensitive to the way in which the classes are numbered. The reason is, correlations between columns of $A$ is independent of $\mathbf{b}$. Indeed, by Theorem \ref{main-thm}, the vector $X-\tilde X$ consisting of the $x_i-\tilde x_i$ is proportional to  correlations between columns of $A$ and as such $X-\tilde X$ is insensitive to changes in  $\mathbf{b}$. Also, if $\mathbf{b}$ changes, then  all the angles between columns of $A$ and  $\mathbf{b}$ will be shifted by a fix amount (the difference of old $\mathbf{b}$ and new $\mathbf{b}$). This shows that our $n\times 3$ matrix is insensitive to the way in which we convert classes to numerical values.

After arriving at the $n\times 3$ matrix, we use a clustering algorithm and cluster our $n\times 3$ matrix into $k$ clusters where $k$ is at most $\textbf{rank}(A)$. Since we do not know the optimal $k$, we take the output feature subset for each $k$ and use a classifier to get an accuracy with respect to that feature subset. Alternatively, our algorithm can take as input an integer $k$ to represent the number of desired features and this way we can just cluster with respect to the input $k$ and return the centroids as the selected subset of features. The final algorithm is presented in Section \ref{algorithm}.

To the best of our knowledge, this is the first work to report on using perturbation theory in feature selection. Specifically, the fact that correlations can be detected via perturbations has not been explored before. As we can see through numerous experiments in Section \ref{experiments}, our method on average chooses smaller number  of features while attaining or exceeding the classification accuracy of other methods. Also, the complexity of our algorithm is dominated by that of computing the SVD of an $m\times n$ matrix which can be done in $O(\min\{mn^2 , m^2n\})$ and even faster as explained in \citep{Holmes}. In particular,  in datasets where we have hundreds of samples and thousands of features ($m^2\leq n$), the complexity of PFS is close to  quadratic. It is also worth noting that our proposed method can be applied to both regression and classification problems. We present some further insights  in Section \ref{Discussion}, and conclude the paper and suggest possible future paths in Section \ref{conclusions}.

\section{Proposed Approach}
Consider the system $AX=\textbf{b}$. Since we want to know the smallest subset of columns of $A$ that we can express $\textbf{b}$ as a linear combination of elements of that subset, we can normalize the columns of $A$. So, we can assume each column of $A$ has length 1.

In real world applications, the system $AX = \textbf{b}$ may not have a solution. In other words, if $b$ is not in the column space of $A$, there is no $X$ such that $AX = \textbf{b}$. Instead, we can  find an $X$ so that the distance between $AX$ and $\textbf{b}$ is as small as possible. That is, we want to minimize the distance $||AX - \textbf{b}||_2$ over all $X$. This minimization problem is known as the method of least squares and its solutions is defined via SVD of $A$. Recall that the SVD of an $m\times n$ matrix $A$ is of the form $A= US V^T $, where $U$ is an $m\times m$ orthogonal matrix, $V$ is an $n\times n$ orthogonal matrix, and $S = \text{diag}(\sigma_1, \ldots, \sigma_r, 0, \ldots, 0 )$ is an $m\times n$ diagonal matrix. Also recall that the Moore-Penrose inverse of $A$ is the $n\times m$ matrix $A^{+}=VS^{-1}U^T$, where $S^{-1} = \text{diag}(\sigma_1^{-1}, \ldots, \sigma_r^{-1}, 0, \ldots, 0 )$.

It is well-known that the least squares solutions can be given in terms of the Moore-Penrose
inverse, see \citep{Golub}.

\begin{thm}[All Least Squares Solutions]\label{ls-solutions} Let $A$ be an $m\times n$ matrix and $\textbf{b}\in \mathbb{R}^m$. Then all the solutions of $\text{min}_{X} ||AX - \textbf{b}||_2$ are of the form $y=A^{+}\textbf{b}+q$, where $q\in \ker(A)$. Furthermore, the unique solution whose 2-norm is the smallest is given by $z=A^{+}\textbf{b}$.
\end{thm}

In our method, each dataset with $m$ samples and $n+1$ features is divided into two matrices: coefficients and constants. Coefficients matrix $A$ involves all the feature values except for the outcome, the constant vector $\textbf{b}$ only contains the classification outcome. In the next section we employ perturbation theory to detect redundant features.

\subsection{Detecting correlations via perturbation}\label{dependency}
To demonstrate how the perturbation can reveal different aspects of features, a synthetic dataset called SynthData is generated with 100 samples and six features based on the following setup:

\begin{align}\nonumber
&\textbf{f}_1 = rand(100), \quad \textbf{f}_2 = rand(100),\\\nonumber
&\textbf{f}_3 = rand(100), \quad \textbf{f}_4 = rand(100),\\\nonumber
&\textbf{f}_5 = 8 \times \textbf{f}_3 + 2 \times \textbf{f}_4,\quad \textbf{f}_6 = 5 \times \textbf{f}_2,\\\nonumber
&\textbf{b} =7\times \textbf{f}_1-3\times \textbf{f}_2+6\times \textbf{f}_3, \nonumber
\end{align}
where $rand(100)$ generates 100 random numbers with uniform probability in the interval $(0, 1)$. So, $D=[A\mid \textbf{b}]$, where $A=[\textbf{f}_1\mid \cdots \mid \textbf{f}_6]$ is an $100\times 6$ matrix. Now let $E$ be a small perturbation of $A$ and solve $AX=\textbf{b}$ and $(A+E)\tilde X=\textbf{b}$ using SVD. We have demonstrated the solutions $X$ and $\tilde X$ as well as their differences in Table \ref{SynthData_perturb}. As we expected, $X$ and $\tilde X$ differ significantly. However, our interest is focused at the last column of Table \ref{SynthData_perturb}, where we have recorded the difference between $X$ and $\tilde X$. 

\begin{table}[H]
\caption{Perturbation of SynthData}
\centering
\begin{tabular}{c c c c}
\hline
& $X$ & $ \tilde{X}$ & $X-\tilde{X}$\\\hline
$x_1$ & 40.8401 & 40.8401 & 2.2115e-05\\
$x_2$ & -8.5981 & -8.5980 & -1.1532e-05\\
$x_3$ & 17.4601 &  -5.9568e+03& -5.9743e+03\\
$x_4$ & -3.7881 & -1.4436e+03 & -1.4398e+03\\
$x_5$ & 16.1273 & 6.1460e+03 & 6.1298e+03\\
$x_6$ & -8.5981 &-8.5980 & -1.8675e-05\\
\hline
\end{tabular}
\label{SynthData_perturb}
\end{table}

Before we state the main theorem, we shall need to recall some facts and definitions which can be found in  \citep{Golub}.

Let $\tilde{A}=A+E$ be a perturbation of $A$. Denote by $\sigma_1\geq \sigma_2\geq \cdots $ and $\sigma'_1\geq \sigma'_2\geq \cdots $ the singular values of $A$ and $\tilde{A}$, respectively. The samllest non-zero singular value of $A$ is denoted by $\sigma_{\text{min}}$ and the greatest of the $\sigma_i$ is  denoted by $\sigma_{\text{max}}$. It is well-known that $||A||_2=\sigma_{\text{max}}$. It has been of great interest to compare the $\sigma_i$ and $\sigma'_i$. In this regard, we use a classical bound on the difference between $\sigma_i$ and $\sigma'_i$ due to Weyl:
\begin{align}\label{weyl}
	|\sigma_i -\sigma'_i| \leq ||E||_2, \quad i=1,2,\cdots
\end{align}

We need to determine the type of perturbations we use. Indeed, we choose $E$ to be a random matrix such that $||E||_2\approx 10^{-s}\sigma_{\text{min}}(A)$, for some $s\geq 0$. We set $s=3$ where our estimates are correct up to a magnitude of $10^{-3}$. We are now ready to prove the main theorem of this paper.

\begin{thm}\label{main-thm} Let $X$ and $\tilde X$ be solutions of $AX=\textbf{b}$ and $(A+E)\tilde X=\textbf{b}$, where $E$ is a small enough perturbation. 
	If a feature $\textbf{f}_i$ is independent of the rest of the features then $|x_i-\tilde x_i|\approx0$. Furthermore, suppose that $S'=\{\textbf{f}_1, \ldots, \textbf{f}_t\}$ is a subset of $S$ such that $\sum_{i=1}^t c_i \textbf{f}_i=0$, for some non-zero $c_i$. If 
	\begin{enumerate}
		\item any subset of $S'$ is linearly independent,
		\item $\textbf{f}_1, \ldots, \textbf{f}_t$ are linearly independent from the rest of features in $S$.
	\end{enumerate}
	Then the vectors 
	$
	\begin{pmatrix}
	c_1\\
	\vdots\\
	c_t
	\end{pmatrix}
	$
	and 
	$
	\begin{pmatrix}
	x_1-\tilde x_1\\
	\vdots\\
	x_t-\tilde x_t
	\end{pmatrix}
	$
	are proportional.
\end{thm}
\Proof
From $AX=\textbf{b}$ and $(A+E)\tilde X=\textbf{b}$, we get $A(X-\tilde X)=E\tilde X$. We claim that $||E\tilde X||\approx 10^{-s}$.
To prove the claim, we consider the SVD of $A+E$ which is of the form $A+E=U\Sigma V^T$. So, $\tilde X=V\Sigma^{-1}U^T b$. Since $U$ and $V$ are orthogonal and for orthogonal matrices we have $|| U\mathbf{v}||_2=||\mathbf{v}||_2$, it follows that 
\begin{align*}
	||\tilde X||_2=||V\Sigma^{-1}U^T \textbf{b}||_2&=||\Sigma^{-1} \textbf{b}||_2\\
	&\leq ||\Sigma^{-1}||_2||\textbf{b}||_2=\frac{1}{\sigma_{\text{min}}(A+E)}\\
	&\leq \frac{1}{-||E||_2+\sigma_{\text{min}}(A)},
\end{align*}
by Equation \eqref{weyl}. Hence, 
\begin{align*}
	||E\tilde X||_2\leq ||E||_2 
	||\tilde X||_2&=\frac{10^{-s}\sigma_{\text{min}}(A)}{-10^{-s}\sigma_{\text{min}}(A)+\sigma_{\text{min}}(A)}\\
	&=\frac{10^{-s}}{1-10^{-s}}=\frac{1}{10^{s}-1}\approx 10^{-s}
\end{align*}

It follows from the claim that 
\begin{align}\label{linear-com}
	(x_1-\tilde x_1)\textbf{f}_1+\cdots +(x_t-\tilde x_t)\textbf{f}_t+\cdots +(x_n-\tilde x_n)\textbf{f}_n\approx 0.
\end{align}
Now, if a feature, say $\textbf{f}_n$, is independent of the rest of features, then it follows from Equation \eqref{linear-com} that $|x_n-\tilde x_n|\approx0$. Suppose now that $S'=\{\textbf{f}_1, \ldots, \textbf{f}_t\}$ is a linearly dependent subset of $S$ such that $\sum_{i=1}^t c_i \textbf{f}_i=0$, for some coefficients $c_1, \ldots, c_t$. 
  Since $\textbf{f}_1, \ldots, \textbf{f}_t$ are linearly independent from the rest of features in $S$, we get 
\begin{align}
	(x_1-\tilde x_1)\textbf{f}_1+\cdots +(x_t-\tilde x_t)\textbf{f}_t\approx 0.\label{depn-rel}
\end{align}
Now, if 
$
\begin{pmatrix}
c_1\\
\vdots\\
c_t
\end{pmatrix}
$
and 
$
\begin{pmatrix}
x_1-\tilde x_1\\
\vdots\\
x_t-\tilde x_t
\end{pmatrix}
$
are not proportional, we can use Equation \eqref{depn-rel} and $\sum_{i=1}^t c_i \textbf{f}_i=0$ to get a dependence relation of a shorter length between the elements of $S'$, which would contradict our  assumption (1). The proof is complete. \qed

Consider now the correlation $\textbf{f}_5 = 8 \times \textbf{f}_3 + 2 \times \textbf{f}_4$ in the SynthData dataset. 
As we mentioned earlier, we normalize the columns of $A$ and replace $A$ with $[\textbf{f}'_1\mid \cdots \mid \textbf{f}'_6]$, where 
$\textbf{f}'_i=\frac{\textbf{f}_i}{|| \textbf{f}'_i||}$. Note that $|| \textbf{f}_3||=5.52, || \textbf{f}_4||=5.33, || \textbf{f}_5||=45.38$. We have
\begin{align*}
\textbf{f}'_5=\frac{\textbf{f}_5}{|| \textbf{f}_5||}=\frac{8 \textbf{f}_3 + 2 \textbf{f}_4}{45.38}
&=0.97 \textbf{f}'_3+0.23\textbf{f}'_4
\end{align*}
So, correlation vector between $\textbf{f}'_3, \textbf{f}'_4, \textbf{f}'_5$ is 
$\begin{bmatrix}
0.97\\0.23\\-1
\end{bmatrix}
$. 
On the other hand, we have 
$\begin{bmatrix}
x_3-\tilde x_3\\
x_4-\tilde x_4\\
x_5-\tilde x_5\\
\end{bmatrix}
=
(-6.1298e+03)
\begin{bmatrix}
0.97\\
0.23\\
-1
\end{bmatrix}
$.
Note that in this example, weights (norms) of $8 \times \textbf{f}_3$ and $\textbf{f}_4$ are very close to each other compared to weight of $2 \times \textbf{f}_4$. In  general,  when a dependence relation exists between a set of features, Theorem \ref{main-thm} along with normalization detect the two features whose weights are closest to each other compared to the others. In particular,  if features $\textbf{f}_i$ and $\textbf{f}_j$ correlate with each other then the differences $\mid x_i-\tilde{x_i}\mid$ and $\mid x_j-\tilde{x_j}\mid$ are almost the same. The converse may not be necessarily true.

We can now consider a column vector whose values are $\mid x_i-\tilde{x_i}\mid$ and use a clustering algorithm to cluster this single column. Clearly, features that correlate with each other fall into the same cluster. However, within a cluster there might be features that do not correlate (but have the same value for $\mid x_i-\tilde{x_i}\mid$). For this reason, we want to further refine the clustering process by computing two more characteristics of data. We shall explain this in the next section.

\subsection{Refining the clustering process}\label{refining clustering}
One way to compare the similarity between vectors is by calculating the angle between them. Features that have smaller angles with the outcome $\textbf{b}$ are informative and predictive. So we construct another column whose values are angles between the $\textbf{f}_i$ and $\textbf{b}$. The angle of each feature with $\textbf{b}$ in SynthData are calculated and shown in the Table \ref{SynthData_angleToB}. 

\begin{table}[H]
\caption{Angle of each feature to $\textbf{b}$ in SynthData}
\centering
\begin{tabular}{c c c c c c c}
\hline
&$\textbf{f}_1$ & $\textbf{f}_2$ & $\textbf{f}_3$ & $\textbf{f}_4$ & $\textbf{f}_5$ & $\textbf{f}_6$\\\hline
$\textbf{b}$ & 37.104 & 112.981 & 47.897 & 87.030 & 48.270 & 112.981\\
\hline
\end{tabular}
\label{SynthData_angleToB}
\end{table}

Our third column in the clustering process is obtained as follows. We remove each feature $\textbf{f}_i$ from the matrix $A$ along with its corresponding coefficient $x_i$ in $X$. Then, the angle of resulting vector $A \setminus \{\textbf{f}_i\} \times X \setminus \{x_i\} =\hat{\textbf{b}}_i$ and the actual outcome $\textbf{b}$ will be considered as a measure of the relevancy for feature $\textbf{f}_i$. 
Note that the closer $\textbf{b}$ and $\hat{\textbf{b}}_i$ are, the less significant the vector $x_i\textbf{f}_i$ is. Applying this process to SynthData is shown in Table \ref{SynthData_angle_to_b}. 

\begin{table}[H]
\caption{Angles of calculated $\hat{\textbf{b}}_i$ to $\textbf{b}$ for SynthData}
\centering
\begin{tabular}{c c c c c c c}
\hline
{Config.} & $\hat{\textbf{b}}_1$ & $\hat{\textbf{b}}_2$ & $\hat{\textbf{b}}_3$ & $\hat{\textbf{b}}_4$ & $\hat{\textbf{b}}_5$ & $\hat{\textbf{b}}_6$ \\\hline
\bf{$\theta$} & 40.390 & 7.748 & 14.574 & 3.507 & 13.330 & 7.748 \\\hline
\end{tabular}
\label{SynthData_angle_to_b}
\end{table}

Now we set up an $n\times 3$ matrix where the first column consists of $|x_i-\tilde x_i|$, the second column is the angles between the  $\textbf{f}_i$'s and  $\mathbf{b}$, and the third column is the angles between the  $\hat{\textbf{b}}_i$'s and  $\mathbf{b}$. Next we use a clustering algorithm to cluster our $n\times 3$ into $k$ clusters. The centroids of clusters will be chosen as our selected  features. Since we do not know the optimal number of clusters, we take the output feature subset for each $k$ and use a classifier to get an accuracy with respect to that feature subset. Alternatively, our algorithm can take as input an integer $k$ to represent the number of desired features and this way we can just cluster with respect to the input $k$ and return the centroids as the selected subset of features. The upper bound for the number of clusters is $\textbf{rank}(A)$, where $\textbf{rank}(A)$ is the numerical rank of $A$.

\subsection{Algorithm}\label{algorithm}
The PFS running time is $t \times (\min(m \times n^2, m^2 \times n)  + k \times (3 \times n \times k))$, where $\min(m \times n^2, m^2 \times n)$ is the complexity of calculating SVD for a $m \times n$ matrix  \citep{Holmes}, and $(3 \times n \times k)$ is the time complexity of the $k$-means clustering algorithm to cluster a dataset of size $n\times 3$ into $k$ clusters. Therefore, the time complexity of PFS is dominated by the complexity of SVD. 

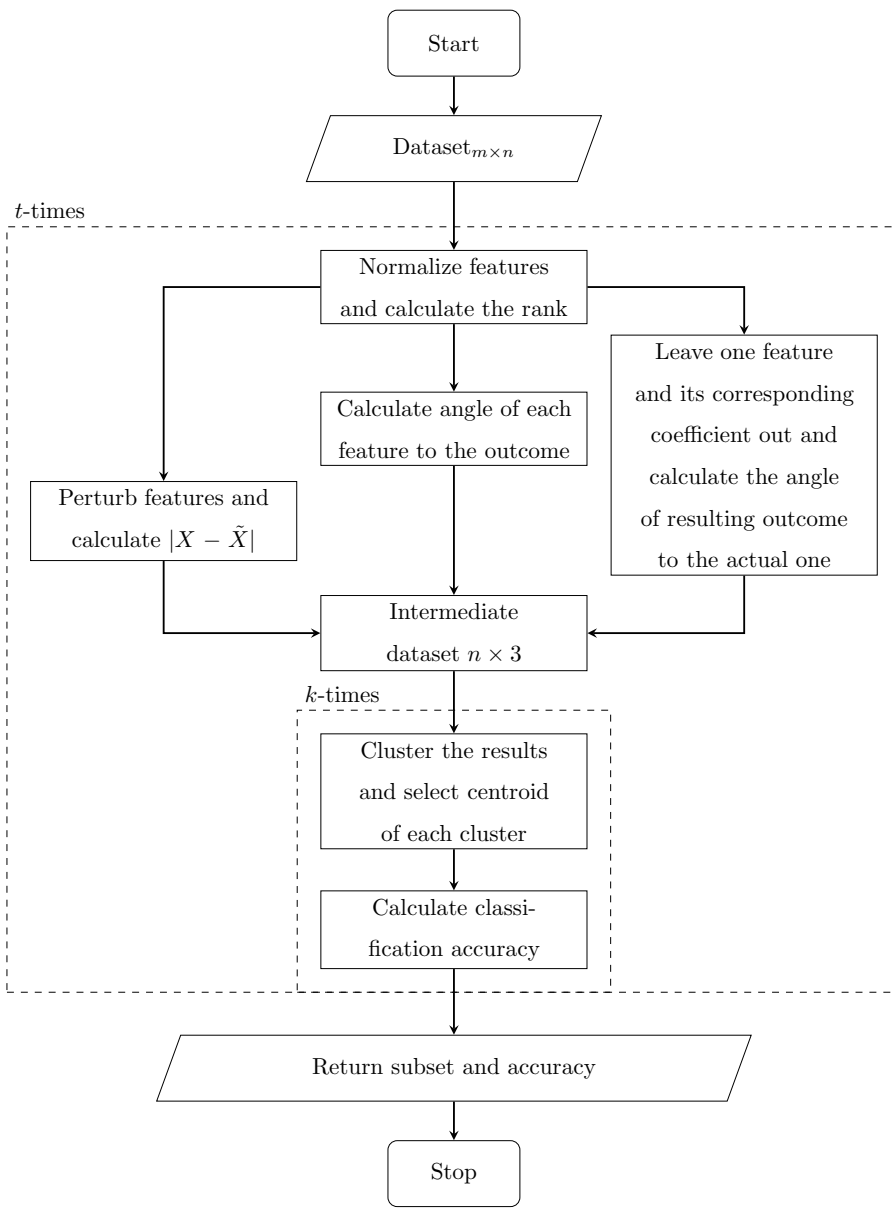
\begin{figure}\label{flowchart}
\centering
\scalebox{.877}{
	\begin{tikzpicture}[node distance=1.6cm ]
	\node (start) [startstop] {Start};
	\node (in1) [io, below of=start] {Dataset$_{m \times n}$};
	\node (pro1) [process, below of=in1, yshift=-.5cm] {Normalize features \\ and calculate the rank};
	\node (pro2) [process, left of=pro1, yshift=-3.55cm, xshift=-2.8cm] {Perturb features and calculate $|X-\tilde X|$};
	
	\node (pro3) [process, below of=pro1, yshift=-.55cm] {Calculate angle of each feature to the outcome};
	\node (pro4) [process, right of=pro1, yshift=-2.55cm, xshift=2.8cm] {Leave one feature and its corresponding coefficient out and calculate the angle of resulting outcome to the actual one};
	
	\node (pro5) [process, below of=pro3, yshift=-1.5cm] {Intermediate dataset ${n \times 3}$};
	\node (pro6) [process, below of=pro5, yshift=-.8cm] {Cluster the results and select centroid of each cluster};
	\node (pro7) [process, below of=pro6, yshift=-.5cm] {Calculate classification accuracy};
	\node (out1) [io, below of=pro7, yshift=-.5cm] {Return subset and accuracy};
	\node (stop) [startstop, below of=out1] {Stop};
	
	\node[container1, fit=(pro1)(pro2)(pro3)(pro4)(pro5)(pro6)(pro7)] (or) {};
	\node at (or.north west) [above right,node distance=0 and 0] {$t$-times};
	
	\node[container1, fit=(pro6) (pro7)] (or) {};
	\node at (or.north west) [above right,node distance=0 and 0] {$k$-times};
	
	\draw [arrow] (start) -- (in1);
	\draw [arrow] (in1) -- (pro1);
	\draw [arrow] (pro1) -| (pro2);
	\draw [arrow] (pro1) -- (pro3);
	\draw [arrow] (pro1) -| (pro4);
	\draw [arrow] (pro2) |- (pro5);
	\draw [arrow] (pro3) -- (pro5);
	\draw [arrow] (pro4) |- (pro5);
	\draw [arrow] (pro5) -- (pro6);
	\draw [arrow] (pro6) -- (pro7);
	\draw [arrow] (pro7) -- (out1);
	\draw [arrow] (out1) -- (stop);
	\end{tikzpicture}
}
\caption{Flowchart of the proposed method}
\end{figure}

Flowchart of PFS is depicted in Figure \ref{flowchart} and is as shown in Algorithm \ref{PFSAlgorithm}. The MATLAB\textsuperscript{\textregistered} implementation of PFS is publicly available on GitHub\footnote{https://github.com/jracp/PerturbationFeatureSelection}.

\begin{algorithm}[!h]
\KwData{$D=[A\mid \textbf{b}]_{m\times n+1}$}
\KwResult{Subset of features and resulting accuracy}
$ACC_{average}$: average accuracy over $t$ runs\;
$ACC_{optimal}$: the optimal  accuracy over $t$ runs \;
$|CLS_{average}|$: average size of subset of features over $t$ runs\;
$|CLS_{optimal}|$: size of the optimal subset of features over $t$ runs\;
Set $t=10, c_l = 10^{6}, c_u = 10^{5}$\;
Normalize columns of $A$ and $\mathbf{b}$ within $[0, 1]$\;
 
 \For{$i = 1$ \KwTo $t$}{
 		$E = (\frac{\max(A)}{c_u} - \frac{\min(A)}{c_l}) \cdot \mathbf{rand} + \min(A)$\;
 		$\tilde{A} = A + E$\;
 		$X = A^{+} \times \textbf{b}$, where $A^{+}$ is the Moore-Penrose inverse of $A$  \;
	 	$\tilde{X} = (\tilde{A})^{+} \times \textbf{b}$\; 
	\For{$j = 1$ \KwTo n}{
		calculate the angle $\theta_j$ between $\textbf{f}_j$ and $\textbf{b}$\;
 		calculate the angle $\gamma_j$ between $A \setminus \{\textbf{f}_j\} \times X \setminus \{x_j\} = \hat{\textbf{b}}_i$ and $\textbf{b}$\;
	}

	\For{$k = 2$ \KwTo $\mathbf{rank}(A)$}{
		Form the $n\times 3$ matrix $[\text{abs}(X-\tilde{X}) \mid \bf{\theta} \mid {\mathbb\gamma} ]$\;
 		cluster this data into $k$ clusters\;
 		find and select centroid features of each cluster\;
 		classify $D$ based on the selected features and return $ACC_{current}$ and $|CLS_{current}|$ \;
	
 	      \If{$ACC_{current} > ACC_{best}$}{
 	    	$ACC_{best} = ACC_{current}$\;
 	    	$|CLS_{best}| = |CLS_{current}|$\;
        	}
        }
    $ACC(j)=ACC_{best}$;
}
Compute and return  $ACC_{average}, |CLS_{average}|, ACC_{optimal}$, and $|CLS_{optimal}|$  based on the vector $ACC$\;
\caption{Perturbation-based Feature Selection}
\label{PFSAlgorithm}
\end{algorithm}

\section{Experimental Results}\label{experiments}
We generate the perturbation matrix $E$ such that the entries of $E$ are randomly chosen in the range  $c_l = 10^{6}$ and  $c_u = 10^{5}$. 
 
Referring to Tran et al. \citep{Tran}, classification accuracy of imbalanced datasets should be calculated using Equation \ref{accuracy}.

\begin{equation}
\frac{1}{s} \sum_{i=1}^{s} \frac{CC_i}{M_i},
\label{accuracy}
\end{equation}

where $s$ is the number of classes in dataset, $CC_i$ is the number of correctly classified instances within class $i$, and $M_i$ is the total number of samples in the class $i$. 

When comparing two feature selection methods, there are three quantities that matter: 1) the accuracy, 2) number of selected features 3) complexity and running time.

We adopt the following formula to compare feature selection methods based on the their accuracy and selected number of features: We quantify the relative effectiveness of a feature selection methods as follows:
\begin{equation}\label{measure}
\frac{\text{classification accuracy}}{\#\, \text{selected features}}.
\end{equation}

Formula \eqref{measure} means that a feature selection method with smaller number of features and higher classification accuracy is  favourable.

All the computations have been done on an ubuntu 14.04 LTS machine with Intel\textsuperscript{\textregistered}Core\texttrademark i5-4570, 24 GB of RAM, using MATLAB\textsuperscript{\textregistered} 9.2.0.556344 (R2017a), R version 3.4.4 (2018-03-15), and Java\texttrademark SE Runtime Environment (build 1.8.0\_151-b12).

\subsection{Comparisons with conventional methods}
In this section, we compare PFS with  Friedman's gradient boosting machine (GBM) \citep{friedman2001greedy}; least absolute shrinkage and selection operator (LASSO) \citep{Tibshirani}; least angle regression (LARS) \citep{Efron}; rescaled linear square regression (RLSR) \citep{chen2017semi} with $k = minSelF$, where $minSelF$ is the minimum number of selected features using GBM, LASSO and LARS; and Hilbert-Schmidt independence criterion least absolute shrinkage and selection operator (HSIC-Lasso) \citep{yamada2014high}. We used gbm package in R \citep{ridgeway2007generalized} for running GBM, and MATLAB\textsuperscript{\textregistered} implementations of LASSO and LARS by Sj{\"{o}}strand \citep{Sjostrand}, RLSR and HSIC-Lasso.

In Section \ref{k-means}, we have used $k$-means to cluster our $n\times 3$ matrix where the upper bound for $k$ is the numerical rank of $A$. To find the best subset, we have experimented with three different classifiers, that is decision tree (DT) \citep{Breiman}, support vector machine (SVM) \citep{allwein2000reducing}, and $k$-nearest neighbour ($k$-NN) \citep{altman1992introduction} in the inner layer. Once we find the $k$ and corresponding subset of features that gives us the best accuracy, we output that subset as the selected features. At the outer layer of our algorithm, we always use DT for classification. To demonstrate a fair and robust result, we run the algorithm 10 times where each time a subset of features is outputted and then classified by DT. The average of accuracies as well as average size of feature subsets are reported. We have demonstrated similar experiments using fuzzy c-means in Section \ref{fuzzy c-means}. 
	
We perform a series of tests on various datasets including, one medical dataset, LSVT Voice  \citep{Tsanas}, one artificial dataset Madelon and six biological datastes -- namely, Colon , Lung, Lymphoma, GLIOMA, Leukemia and ALLAML   -- have been selected from ASU dataset repository  \citep{feature} and UCI repository of machine learning  \citep{Lichman}. The specifications of all datasets are given in Table \ref{data}. 

\begin{table}[]
\caption{Dataset Specifications}
\centering
\begin{tabular}{l c c }
\hline
{Dataset} & {Samples} & {Features}\\\hline
LSVT Voice & 126 & 310\\
Madelon & 2000 & 500\\
Colon & 62 & 2000\\
Lung & 203 & 3312\\
Lymphoma & 96 & 4026\\
GLIOMA & 50 & 4434\\
Leukemia & 72 & 7070\\
ALLAML & 72 & 7129\\
\hline
\end{tabular}
\label{data}
\end{table}

Note that for the experiments in this section, the decision tree classifier is applied with  MATLAB\textsuperscript{\textregistered}, using 70\% of the data for training and 30\% for testing and validating. This set up is applied to all methods including GBM, LASSO, LARS, RLSR, HSIC-Lasso, and PFS. Since PFS uses a clustering algorithm, the selected subset of features in PFS can change each run. So, we run PFS 10 times on randomly shuffled data where testing and trainings sets vary accordingly in each run. 

\subsubsection{Evaluation results using $k$-means}\label{k-means}
In this section, we  use $k$-means to cluster our $n\times 3$ matrix where the upper bound for $k$ is the numerical rank of $A$. To find the best subset, we have experimented with three different classifiers, that is  DT, SVM  and $k$NN in the inner layer. Once we find the $k$ and corresponding subset of features that gives us the best accuracy, we output that subset as the selected features. At the outer layer of our algorithm, we always use DT for classification for all the methods. 

In Tables \ref{results_kmeans_features} and \ref{results_kmeans_classification}, we have reported the selected number of feature and classification accuracies, respectively. Note that PFS-DT, PFS-SVM,  and PFS-$k$NN mean that we have used DT, SVM, and $k$NN as the inner classifier in PFS, respectively. In all the methods we have used DT to report the classification accuracy.

\begin{table}[H]
	\centering
	\caption{Number of selected features using GBM, LASSO, LARS, RLSR, HSIC-Lasso, PFS based on decision tree classifier (PFS-DT), PFS based on support vector machine classifier (PFS-SVM) and PFS based on $k$-nearest neighbour classifier (PFS-$k$NN). For each version of PFS the mean of the number of selected features in 10 run is reported in subscript.}
	\centering
	\tabcolsep=0.02cm
	\begin{tabular}{l c c c c c c c c c c c}
		\hline
		\multirow{2}{*}{{Dataset}} & \multicolumn{8}{c}{{Number of selected features}}\\
		& {\small{GBM}} & {\small{LASSO}} & {\small{LARS}}& {\small{RLSR}} & {\small{HSIC-Lasso}} &{\small{PFS-DT}}&{\small{PFS-SVM}} & {\small{PFS-$k$NN}}\\\hline
		LSVT Voice &239&126&125&125&12&13$_{45.30}$&87$_{111.90}$&30$_{94.60}$\\
		Madelon &467&89&89&89&|&34$_{100.80}$&6$_{24.80}$&25$_{64.60}$\\
		Colon &656&62&61&61&9&7$_{29.80}$&22$_{39.30}$&18$_{30.60}$\\
		Lung &1503&203&202&202&134&34$_{105.00}$&28$_{100.00}$&58$_{131.20}$\\
		Lymphoma &1491&96&95&95&181&36$_{51.80}$&23$_{44.80}$&42$_{75.50}$\\
		GLIOMA &535&50&49&49&17&7$_{25.60}$&17$_{36.50}$&28$_{37.50}$\\
		Leukemia &1053&72&71&71&17&6$_{46.10}$&15$_{41.00}$&24$_{49.00}$\\
		ALLAML &1200&72&71&71&8&15$_{41.20}$&24$_{53.40}$&8$_{43.00}$\\
		\hline
	\end{tabular}
	
	\label{results_kmeans_features}
\end{table}

To demonstrate a fair and robust result, we run our algorithm 10 times where each time the dataset is  randomly shuffled and a subset of features is outputted.  The average of accuracies as well as average size of feature subsets are reported. Also, we use Formula \ref{measure} to find the optimal accuracy and subset of features amongst the 10 run. In columns corresponding to PFS-DT, PFS-SVM, and PFS-$k$NN, the optimal number of features and optimal classification accuracy with respect to Formula \ref{measure} are shown in the superscript whereas the average number of features and    average of classification accuracies  are shown in the subscript. 

\begin{table}[H]
	\small
	\centering
	\caption{Classification accuracies of GBM, LASSO, LARS, RLSR, HSIC-Lasso, PFS based on decision tree classifier (PFS-DT), PFS based on support vector machine classifier (PFS-SVM) and PFS based on $k$-nearest neighbour classifier (PFS-$k$NN). For each version of PFS the mean of the resulting classification accuracies in 10 run  is reported in subscript.}
	\centering
	\tabcolsep=0.03cm
	\begin{tabular}{l c c c c c c c c c c c}
		\hline
		\multirow{2}{*}{{Dataset}} & \multicolumn{8}{c}{{Classification Accuracy}}\\
		& {GBM} & {LASSO} & {LARS}& {RLSR} & {HSIC-Lasso} &{PFS-DT}&{PFS-SVM} & {PFS-$k$NN}\\\hline
		LSVT Voice &73.68&73.68&72.14&63.16&78.94&83.97$_{85.26}$&60.00$_{64.46}$&84.28$_{86.86}$\\
		Madelon &77.67&53.16&62.00&49.34&|&76.18$_{81.45}$&62.15$_{61.62}$&83.67$_{81.97}$\\
		Colon &78.95&83.33&79.49&68.42&84.21&100.00$_{91.58}$&89.20$_{92.61}$&84.66$_{89.20}$\\
		Lung &75.41&51.17&63.58&75.41&83.60&96.20$_{94.10}$&100.00$_{99.95}$&100.00$_{99.84}$\\
		Lymphoma &62.07&39.21&32.19&60.71&51.72&64.65$_{55.93}$&61.11$_{62.41}$&66.67$_{69.94}$\\
		GLIOMA &60.00&52.50&53.75&53.33&80.00&85.42$_{79.33}$&95.00$_{90.08}$&95.00$_{85.58}$\\
		Leukemia &95.46&96.88&96.88&95.46&100.00&96.88$_{95.45}$&97.06$_{99.71}$&97.06$_{98.23}$\\
		ALLAML &90.91&90.83&90.83&62.38&90.90&93.33$_{89.09}$&93.33$_{96.29}$&85.71$_{90.95}$\\
		\hline
	\end{tabular}
	\label{results_kmeans_classification}
\end{table}

We can see from Table \ref{results_kmeans_classification} that, over all, the classification accuracies of PFS-based methods are favourable to the other methods and only HSIC-Lasso is sometimes attaining similar accuracies. On the other hand, HSIC-Lasso chooses less number of features on average compared to PFS-based methods. We remark that the number of features in PFS depends on the upper bound we set for the number of clusters when we cluster our intermediate $n\times 3$ matrix. We have taken $\mathbf{rank}(A)$ as an upper bound but this bound is just a crude estimate and in the next phases of this project we shall improve this bound. Hence, it is possible to still decrease the average number of features in PFS.
 
We can also observe from Table \ref{results_kmeans_classification}, that when $k$NN is used as the inner classifier, the average classification accuracies are slightly  better than when DT or SVM are used. In contrast, the average number of features  are slightly  lower  when DT is used as the inner classifier. 
 
\subsubsection{Evaluation results using fuzzy $c$-means}\label{fuzzy c-means}
To investigate the affect of clustering method, we have also experimented with  fuzzy $c$-means clustering algorithm for which, the results are shown in Table \ref{results_cmean}. We can also observe from Table \ref{results_cmean} that all in all there is very little difference in    average classification accuracies regardless of which classifier is used. In contrast, the average number of features  are slightly  lower  when DT is used as the inner classifier.

\begin{table}[!h]
\centering
\caption{The number of selected features and the resulting classification accuracies using fuzzy $c$-means version of PFS based on decision tree classifier (PFS-DT), PFS based on support vector machine classifier (PFS-SVM) and PFS based on $k$-nearest neighbour classifier (PFS-$k$NN). For each version of PFS the mean of the number of selected features and the mean of the resulting classification accuracies is reported in subscript.}
\centering
\tabcolsep=0.07cm
\begin{tabular}{l c c c c c c c}
\hline
\multirow{2}{*}{{Dataset}} & \multicolumn{3}{c}{{Number of selected features}} & \multicolumn{3}{c}{{Classification Accuracy}}\\
 &{PFS-DT}&{PFS-SVM} & {PFS-$k$NN}  &{PFS-DT}&{PFS-SVM} & {PFS-$k$NN}\\
\hline 
LSVT Voice &15$_{55.70}$&2$_{87.70}$&67$_{86.40}$& 89.74$_{82.43}$&50.00$_{56.00}$&81.07$_{86.11}$ \\
Madelon & 19$_{154.80}$&15$_{175.80}$&78$_{127.80}$&75.35$_{81.27}$&62.48$_{61.45}$&79.66$_{80.42}$ \\
Colon & 11$_{33.10}$&13$_{29.80}$&13$_{33.70}$&86.67$_{89.15}$&90.91$_{89.77}$&89.20$_{88.86}$ \\
Lung & 53$_{93.00}$&66$_{126.50}$&63$_{133.90}$&95.79$_{90.88}$&99.47$_{98.96}$&$100.00_{98.42}$ \\
Lymphoma &59$_{53.20}$&13$_{37.80}$&58$_{73.40}$&69.23$_{53.18}$&63.58$_{62.28}$&76.54$_{71.05}$ \\
GLIOMA &5$_{30.40}$&15$_{31.50}$&17$_{31.60}$&89.58$_{79.00}$&90.00$_{88.67}$&86.67$_{87.25}$ \\
Leukemia &7$_{31.60}$&18$_{42.60}$&17$_{44.70}$&100.00$_{97.65}$&94.12$_{97.35}$&94.12$_{96.06}$ \\
ALLAML &27$_{44.60}$&32$_{58.10}$&8$_{51.60}$&86.09$_{89.81}$&82.86$_{86.90}$&90.00$_{87.29}$ \\
\hline
\end{tabular}
\label{results_cmean}
\end{table}

\subsubsection{A quantified measure}
In Sections \ref{fuzzy c-means} and \ref{k-means}, we have used each of $k$-means and fuzzy $c$-means as our clustering algorithm. It seems that using fuzzy $c$-means, our method in general chooses more features. To present and amalgamate the results of Tables \ref{results_kmeans_features},
\ref{results_kmeans_classification}, and \ref{results_cmean}, we  apply Formula \ref{measure} using average classification accuracy and average number of features to obtain a  comparison in Table \ref{results_measure} between  $k$-means and fuzzy $c$-means. We can conclude that based on the measure given by Formula \ref{measure}, our algorithm has a better performance when $k$-means is used for clustering. 

\begin{table}[!h]
	\centering
	\caption{The resulting measure calculated using Equation \ref{measure} for $k$-means and $c$-means versions of PFS based on decision tree classifier (PFS-DT), PFS based on support vector machine classifier (PFS-SVM) and PFS based on $k$-nearest neighbour classifier (PFS-$k$NN).}
	\centering
	\tabcolsep=0.1cm
	\begin{tabular}{l c c c c c c c}
		\hline
		\multirow{2}{*}{{Dataset}} & \multicolumn{3}{c}{{$k$-means}} & \multicolumn{3}{c}{{$c$-means}}\\
		&{PFS-DT}&{PFS-SVM} & {PFS-$k$NN}  &{PFS-DT}&{PFS-SVM} & {PFS-$k$NN}\\
		\hline 
		LSVT Voice &   1.88&   0.57&   0.91&   1.47&   0.64&   1.00\\
		Madelon &   0.81&   2.54&   1.26&   0.52&   0.34&  0.62\\
		Colon &   3.95&   2.35&   2.96&   2.69&   3.06&  2.66\\
		Lung &   0.89&   0.99&   0.75&   0.96&   0.77&   0.73\\
		Lymphoma &   1.03&   1.40&   0.92&   1.00&   1.67&   0.97\\
		GLIOMA &   3.16&   2.50&   2.29&   2.63&   2.83&   2.80\\
		Leukemia &   4.52&   2.41&   2.00&   3.12&   2.30&   2.18\\
		ALLAML &   2.17&  1.81&   2.09&   2.02&   1.48&   1.70\\
		\hline
	\end{tabular}
	\label{results_measure}
\end{table}

\subsection{Comparison with methods based on SVM \&  optimization}\label{cancer}
A recent paper by Ghaddar and Naoum-Sawaya  \citep{Ghaddar} proposed a feature selection method using support vector machines (FS-SVM) for binary-class datasets, in which, a pre-defined percentage of features is selected through adjusting $l_1-$norm of the classifier. 

Ghaddar et al. applied their method to a set of cancer datasets (\# of samples $\times$ \# of features) -- namely, Leukemia (72 $\times$ 7130), Lung cancer (139 $\times$ 1000), Prostate cancer (102 $\times$ 12,601) -- adopted from Cancer Program at Broad Institute \footnote{http://www.broad.mit.edu/cgi-bin/cancer/datasets.cgi} (different form those in Table \ref{data}). For each dataset, a subset of positive and negative classes have been selected for training and testing purposes (see Table \ref{ghaddar_config}).

\begin{table}[H]
\centering
\caption{Number of samples of each class for each dataset in FS-SVM}
\centering
\begin{tabular}{l c c c c}
\hline
\multirow{2}{*}{{Dataset}} & \multicolumn{2}{c}{{Train}} & \multicolumn{2}{c}{{Test}}\\
& Class 1 & Class 2 & Class 1 & Class 2\\\hline
Leukemia & 24 & 13 & 23 & 12 \\
Lung     & 9 & 70 & 8 & 69\\
Prostate  & 25 & 26 & 25 & 26\\
\hline
\end{tabular}
\label{ghaddar_config}
\end{table}

We have used PFS with DT as the inner classifier and followed  the same setup to compare PFS-DT with the method proposed in \citep{Ghaddar}. To get unbiased results, we run PFS-DT 10 times where each time we shuffled and constructed test and  train datasets based   on the configuration in Table \ref{ghaddar_config}. The optimal and average results are reported in Table \ref{results_ghaddar}.

In order to find the highest classification accuracy, the authors in \citep{Ghaddar} have applied their method FS-SVM and limited the selected subset of features to range from 2\% to 20\% of total number of features. In turn, the running time of FS-SVM is very high.

\begin{table}[H]
\centering
\caption{Comparison of  PFS based on decision tree classifier (PFS-DT) and FS-SVM}
\centering
\begin{tabular}{l c c c c}
\hline
\multirow{2}{*}{{Dataset}} & \multicolumn{2}{c}{{Number of selected features}} & \multicolumn{2}{c}{{Classification Accuracy}}\\
 & {FS-SVM} & {PFS-DT}& {FS-SVM} & {PFS-DT}\\\hline
Leukemia &142& {24$_{20.4}$} & 80.00 & {85.15$_{77.34}$}\\
Lung     &20 & {3$_{29.90}$} & 97.00 & {100.00$_{99.28}$}\\
Prostate  &252& {29$_{37.40}$} & 86.00 & {88.23$_{87.44}$}\\
\hline
\end{tabular}
\label{results_ghaddar}
\end{table}

\section{Discussions}\label{Discussion}
The upper bound for the number of clusters in Algorithm 1 is the numerical rank of matrix $A$, which infers about the largest number of independent features. There exists various clustering algorithms and one way to improve the proposed method is to cluster the generated characteristics dataset more efficiently. Of course, the number of clusters in PFS can be set manually which adds a great flexibility in selecting a certain number of features. It is worth noting that some of the clusters that represent irrelevant features can be excluded right away before starting the clustering process. Irrelevant features can be detected by their corresponding coefficients in the solution of the least squares problem. 

Since  $k$-means and fuzzy $c$-means clustering method choose the initial centroids randomly, the final outcome of PFS could be  different per run, which introduces a valid concern of non-reproducibility of the results. To remedy this, the proposed algorithm has iterated $t$-times to provide more robust and reproducible results. An alternative approach is to use a deterministic clustering algorithm which we shall examine in the future.

The complexity of our proposed method is dominated by the complexity of calculating SVD. 

\section{Conclusions and future work}\label{conclusions}
In this paper, we proposed a novel  feature selection method. We divide  a dataset $D$  into a  matrix $A$ consisting of    features and the vector $\textbf{b}$ of  the classification outcome, hence $D=[A\mid \textbf{b}]$. We solve the least squares problem $\text{min}_X ||AX-\textbf{b}||_2$ using the singular decomposition of $A$. We have proved and demonstrated how perturbation theory can be used to detect correlations between features. Through this process, irrelevant features can be identified and filtered out at the very first stages of the algorithm. The main ingredient of our approach is perturbation theory and experimental results show how powerful this method is to detect and remove correlations.  We have compared our method with several other methods and it is shown that PFS always chooses a fraction of the number of features selected by other methods. Furthermore, we believe PFS is robust against noise. A noisy data can be viewed as a perturbed system. So we can consider a system of the form $\tilde AX=\tilde{\textbf{b}}$ and apply Theorem \ref{main-thm}.  We shall investigate the noise-robustness of PFS  in future work.

We compared the results from our method with famous LASSO and LARS methods and their descendants RLSR and HSIC-Lasso, as well as, GBM against several datasets.  Moreover, we compared our method with the recently proposed  method based on optimizing the support vector machines (FS-SVM)  \citep{Ghaddar}. The overall performance of PFS in terms of the number of selected features and resulting classification accuracies shows its applicability and effectiveness compared to conventional and recent feature selection methods.

The advantage of the proposed method is its modularity. It can be seen as a framework for future feature selection methods, in which different characteristics of feature are extracted using a set of measures. Then, the results are grouped using a user-specified clustering method. Finally, each cluster is evaluated by an arbitrary classifier and the best subset is selected either based on the size of the selected subset or resulting classification accuracy or a combination of both, as suggested in Equation \ref{measure}.

In a future work, we shall also investigate the effect of using different parametric and non-parametric clustering methods to compare the results and decrease the complexity of PFS. Also, we are looking at designing a version of the PFS applicable to gene datasets through a multi-stage process.

\section*{Acknowledgements}The research of the second author  was supported by NSERC of Canada under grant \# RGPIN 418201. The authors would like to thank the anonymous reviewers for valuable comments and feedback that helped with the exposition and clarity of results.

\bibliography{mybib}

\end{document}